\documentclass[runningheads]{llncs}
\usepackage[T1]{fontenc}
\usepackage{graphicx}
\usepackage{subfigure}
\usepackage{xcolor}
\usepackage{adjustbox}
\usepackage{hyperref}

\usepackage{multirow}
\usepackage[flushleft]{threeparttable}

\begin{document}
\title{Investigating Behavior Cloning from Few Demonstrations for Autonomous Driving based on Bird's-Eye View in Simulated Cities\thanks{Accepted manuscript of the paper published in \emph{Intelligent Systems: 34th Brazilian Conference, BRACIS 2024, Bel\'em, Brazil, November 17--21, 2024, Proceedings}, Lecture Notes in Computer Science (LNAI), vol.~15413, Springer, 2025, pp.~155--168. This version of the contribution has been accepted for publication, after peer review, but is not the Version of Record and does not reflect post-acceptance improvements, or any corrections. The Version of Record is available online at \url{https://doi.org/10.1007/978-3-031-79032-4_11}. Use of this Accepted Version is subject to the publisher's Accepted Manuscript terms of use: \url{https://www.springernature.com/gp/open-research/policies/accepted-manuscript-terms}.}}
\titlerunning{Behavior Cloning from Few Demonstrations for Autonomous Driving}
\author{
Eric Aislan Antonelo\inst{1} \and
Gustavo Claudio Karl Couto\inst{1} \and
Christian Möller\inst{2} \and
Pedro Henrique Fernandes\inst{1}
}
\authorrunning{E. A. Antonelo et al.}
\institute{
Automation and Systems Engineering Department, Federal University of Santa Catarina, Florianópolis, Brazil\\ 
\email{eric.antonelo@ufsc.br, 	gustavo.karl.couto@posgrad.ufsc.br}\\
\and
Faculty of Science and Engineering, Information Technology\\
Åbo Akademi University, Finland \\
\email{christian.moller@abo.fi}
}
\maketitle              %
\begin{abstract}
This paper investigates the use of Behavior Cloning (BC) for autonomous driving from a bird’s-eye view (BEV) perspective in simulated urban environments. 
BC uses supervised learning to mimic expert driving behaviors.
Previous works have applied BC in the CARLA simulator but did not fully address the challenges of traffic light compliance. 
Our approach enhances BC by integrating a kernel density estimator to adjust training sample weights based on action density, thereby improving the learning of rare but critical actions such as stopping at red lights and accelerating at green lights, specially in scenarios of scarce number of expert demonstrations. Using BEV inputs, which provide an abstract top-down view of the driving environment, our method simplifies the policy learning process. The trained convolutional neural network (CNN) outputs steering and acceleration actions based on these BEV inputs and additional state variables. Experimental results in the CARLA simulator demonstrate that our weighted BC method significantly improves driving performance, achieving higher route completion compared to standard BC. 
This weighted approach proved to be crucial in learning correct driving behaviors, particularly in test environments not encountered during training, highlighting its potential for enhancing autonomous vehicle navigation. 

\end{abstract}
\section{Introduction}

Autonomous driving has traditionally been implemented using separate modules for perception, planning, and control, organized in a sequential pipeline \cite{juniordarpa,env_perception,surveypipeline,Parizotto2021}. However, learning-based approaches have been gaining popularity as they attempt to address the complexities of autonomous driving in various scenarios, whether in simulation or real-world environments.

Most of these learning-based approaches rely on Behavior Cloning (BC), which employs supervised learning on a collection of expert demonstrations gathered offline \cite{nvidiabc,videobc,chauffernet,codevilla2018endtoend,codevilla2019exploring}. For example, a human driver generates a set of input (observations) and corresponding desired output pairs.

Codevilla \cite{codevilla2018endtoend,codevilla2019exploring} employed BC for autonomous driving in the CARLA simulator. 
He collected a large dataset of human driving and augmented it using image processing techniques to train end-to-end policies conditioned on the desired route. 
Further improvements in BC were introduced in \cite{ral_prob_bc}, where a large deep ResNet network was used for feature extraction. 
This approach combined data from cameras, LiDAR, and radar to create comprehensive feature maps.
The work also incorporated a probabilistic motion planner to account for uncertainties in the trajectory.

Couto \cite{Couto2021} has shown results on behavior cloning on the CARLA simulator, but the agent did not have to respect traffic lights. In this work, we investigate an approach where the agent is able to navigate the city while respecting traffic lights, i.e., breaking when the light is red, and crossing the traffic light when it is green. 
Only relying on regular BC to train an agent is not enough in a scenario with limited amount of training routes and where the agent needs to break at the red light and accelerate from stillness when light turns to green. To solve that, we employ a kernel density estimator \cite{Silverman2018density} to estimate the actions density from the expert dataset. Then, the training samples are weighted in the BC loss function individually by setting the weight equal to the inverse of the actions' density. In this way, we can account for the underrepresented actions in the expert set, i.e., those happening at traffic light transitions. This weighting is needed because we are considering only eight routes as training data.

Instead of learning directly from raw camera pixels as in \cite{Couto2021,cirl2018}, the agent in this work has structured image inputs that are more abstract and meaningful to the navigation task.
To decouple feature representation from policy learning \cite{resnet_rl,Couto2023} has its advantages as potentially easing up the learning task. 
In our context, this decoupling is accomplished by using a mid-level representation of the vehicle's raw camera images. 
For example, Bird's-Eye View (BEV) representations of the road ahead have been used as mid-level inputs for trajectory generation and motor control networks in \cite{chauffernet} and \cite{roach}, respectively.
In BEV representation, the camera scene is projected onto a top-down view, segmenting areas of interest, like vehicle lanes and sidewalks, into different colors \cite{env_perception}. 
By using BEV as input for the control policy, the learning process can focus on the navigation problem, leveraging a more abstract and simplified view of the road ahead. The policy here is represented by convolutional neural networks, mapping the BEV image to control actuators such as steering and acceleration.

Experimental results in the CARLA simulator demonstrate that our weighted BC method significantly improves driving performance, achieving higher route completion compared to standard BC. 
This weighted approach proved to be essential in learning correct driving behaviors, particularly in test environments not encountered during training.

In Section~\ref{sec:methods}, we present
the bird's-eye view representation employed as input to the convolutional neural network, the behavior cloning loss for the agent, and the kernel density estimation used to generate the weights per sample for the modified loss function. The agent architecture is presented in Section ~\ref{sec:agent}.
Finally, the experiments and conclusions are shown in Sections \ref{sec:experiments} and \ref{sec:conclusion}.

\section{Methods}
\label{sec:methods}

\subsection{Bird's-Eye View - BEV representation}
\begin{figure*}[thpb]
  \subfigure[Frontal Cameras]{
    \label{fig:cameras}  
    {\includegraphics[scale=0.42]{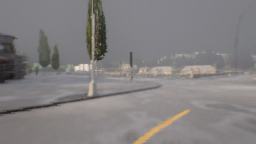}}
    {\includegraphics[scale=0.42]{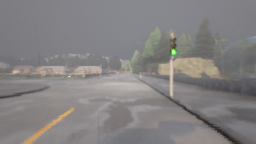}}
    {\includegraphics[scale=0.42]{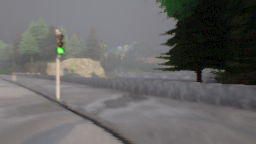}}
  }
  \subfigure[Bird's-Eye View (BEV) channels: route, drivable area, and lane boundaries]{
    {\includegraphics[scale=0.39]{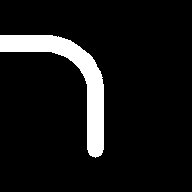}}
    \quad
    {\includegraphics[scale=0.39]{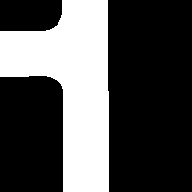}}
    \quad
    {\includegraphics[scale=0.39]{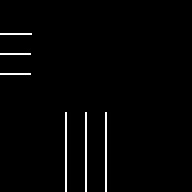}}
    \qquad \qquad \qquad
    \qquad
    \label{fig:bev}
  }
  \subfigure[BEV's traffic lights channels at timesteps (-1, -9, -16)]{
    {\includegraphics[scale=0.39]{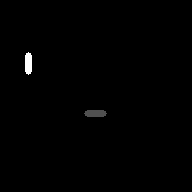}}
    \quad
    {\includegraphics[scale=0.39]{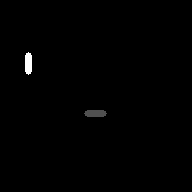}}
    \quad
    {\includegraphics[scale=0.39]{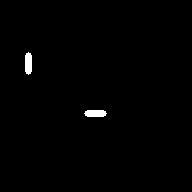}}
    \label{fig:bev_lights}
    }
\subfigure[final BEV image]{ 
    {\includegraphics[scale=0.39]{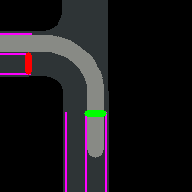}}
    \label{fig:bev_final}
  }
  \caption{a) Images from three frontal cameras positioned on the left, center, and right side of the vehicle. These images were taken after the initial interactions of the agent in the CARLA simulation environment. Each camera captures a 256$\times$144 RGB image.
  The corresponding bird's-eye view channels that our agent uses are shown in (b), (c) and (d), computed at the same moment as shown in (a).
b) Three BEV 192$\times$192 channels that represent, from left to right, the desired route, drivable area, and lane boundaries. 
c) BEV's traffic lights channels at timesteps (-1, -9, -16), i.e., representing the past states of the traffic light channel (grey color is red light off; white color means red ligh on).
d)
The final image displays all three channels combined in different colors.
  }
  \label{fig:cameras_all}
\end{figure*}

The bird's-eye view of a vehicle represents its position and movement in a top-down coordinate system 
\cite{chauffernet}. The vehicle's location, heading, and speed are represented by $p_t$, $\theta_t$, and $s_t$ respectively. 
The top-down view is defined so that the agent's starting position is always at a fixed point within an image (the center of it). 
Furthermore, it is represented by a set of images of size $W \times H$ pixels, at a ground sampling resolution of $\phi$ meters/pixel. 
The BEV of the environment moves as the vehicle moves, allowing the agent to see a fixed range of meters in front of it. For instance, the BEV representation for the vehicle whose three frontal cameras are shown in Fig.~\ref{fig:cameras} is given in Fig.~\ref{fig:bev}, where the desired route, drivable area and lane boundaries are shown as three BEV channels. Besides, other three BEV channels for the traffic lights representing the state of the traffic lights (green or red) at different timesteps are shown in Fig.~\ref{fig:bev_lights}. Fig.~\ref{fig:bev_final}
shows the integrated six-channel BEV image used as input to the agent.

\subsection{Behavior Cloning (BC)}
Behavior cloning is a supervised learning approach used in imitation learning, where the goal is to learn a policy that mimics the behavior of an expert. 
In this work, we want to learn a policy for an autonomous vehicle that outputs two actions (acceleration, steering) based on a Bird-Eye-View (BEV) representation of the environment and additional scalar state variables as input. For this reason we need to train a CNN to output these two continuous actions based on input from the BEV and the state variables. 

The loss function for behavior cloning is typically based on the difference between the actions predicted by the learned policy and the actions taken by the expert in the training data.
Let:
\( \pi_{\theta}(s) \) be the policy parameterized by \(\theta\), which maps a state \( s \) to an action \( a \);
and \( \mathcal{D} = \{(s_i, a_i)\}_{i=1}^N \) be the dataset of \(N\) state-action pairs collected from the expert, where \( s_i \) is a state and \( a_i \) is the corresponding action taken by the expert.

Although the behavior cloning loss function can be defined as the mean squared error (MSE) between the predicted actions and the expert actions (or the cross-entropy loss if the actions are discrete),
in this work, we model a stochastic policy and, for that reason, the policy is represented by
a Beta distribution. 
The Beta distribution is suitable for modeling variables that are bounded within an interval, such as acceleration and steering actions within \([-1, 1]\). 

Given the actions \(a_i = (a_{i}^\mathrm{acc}, a_{i}^\mathrm{steer})\) and the policy \(\pi_{\theta}(s_i) = (\alpha_{i}^\mathrm{acc}, \beta_{i}^\mathrm{acc}, \alpha_{i}^\mathrm{steer},\)
\(\beta_{i}^\mathrm{steer})\), 
where \(\alpha\) and \(\beta\) are the shape parameters of the Beta distribution, the loss function will be based on the negative log-likelihood of the Beta distribution.

The probability density function of the Beta distribution \cite{Gelman1995bayesian} is:
\[ f(x; \alpha, \beta) = \frac{x^{\alpha-1} (1-x)^{\beta-1}}{B(\alpha, \beta)}, \]
where \(B(\alpha, \beta)\) is the Beta function, a normalization constant to ensure that the total probability is 1.
To transform actions in \([-1, 1]\) to \([0, 1]\), we use:
\[ a_i' = \frac{a_i + 1}{2} \]

The negative log-likelihood loss for the Beta distribution can then be written as:

\begin{equation}
\mathcal{L}(\theta) = -\frac{1}{N} \sum_{i=1}^N \left[ \log f\left( \frac{a_{i}^\mathrm{acc} + 1}{2}; \alpha_{i}^\mathrm{acc}, \beta_{i}^\mathrm{acc} \right) + \log f\left( \frac{a_{i}^\mathrm{steer} + 1}{2}; \alpha_{i}^\mathrm{steer}, \beta_{i}^\mathrm{steer} \right) \right] 
    \label{eq:bcloss_car}
\end{equation}

This loss function accounts for the probabilistic nature of the Beta distribution, ensuring that the model learns to predict parameters (\(\alpha, \beta\)) for the distribution of actions rather than the actions directly. This approach allows for modeling uncertainty and variability in the predicted actions.

\subsection{Kernel Density Estimation}

A Kernel Density Estimator (KDE) is a non-parametric way to estimate the probability density function (PDF) of a random variable.
For each data point in the sample, the kernel function is centered at that point. The KDE is then computed by summing up all these kernel functions and normalizing by the number of data points. 
Mathematically, the KDE for a point \(x\) is given by:

\begin{equation}
   \hat{g}(x) = \frac{1}{N h} \sum_{j=1}^{N} K \left( \frac{x - x_j}{h} \right),
    \label{eq:kde}
\end{equation}
where \( \hat{g}(x) \) is the estimated density at point \( x \), \( N \) is the number of data points, \( h \) is the bandwidth, \( x_j \) are the data points, and \( K \) is the kernel function.

The Gaussian kernel, often used in KDE, is a type of kernel function that assigns weights to data points based on their distance from the point where the density is being estimated. 
The Gaussian kernel is defined by the standard normal distribution, and it ensures that closer points have a higher influence on the density estimate than points further away.
The mathematical form of the Gaussian kernel \( K \) is:

\[ K(u) = \frac{1}{\sqrt{2\pi}} e^{-\frac{u^2}{2}}, \]
where \( u \) is the standardized distance between the point where the density is being estimated and the data point, scaled by the bandwidth \( h \).

In this work, the KDE is used to weigh the training samples in the BC loss in (\ref{eq:bcloss_car}).
By using the inverse of the kernel density estimation of the actions as weights, we give more importance to less frequent actions. This approach can help the model learn better on underrepresented parts of the action space.

The resulting weighted BC loss function can be written as:

\begin{equation}
 \mathcal{L}(\theta) = -\frac{1}{N} \sum_{i=1}^N w_i \left[ \log f\left( \frac{a_{i}^\mathrm{ac} + 1}{2}; \alpha_{i}^\mathrm{ac}, \beta_{i}^\mathrm{ac} \right) + \log f\left( \frac{a_{i}^\mathrm{steer} + 1}{2}; \alpha_{i}^\mathrm{steer}, \beta_{i}^\mathrm{steer} \right) \right] 
    \label{eq:weighted_loss}
\end{equation}
where:
\( w_i = \frac{1}{\hat{g}(a_i)} \);
and \( \hat{g}(a_i) \) is the kernel density estimation evaluated at the action \( a_i \).
The KDE is estimated for the actions in the training set. Then the weights $w_i$ are calculated as the inverse of the KDE for each action.

This approach ensures that the loss function accounts for the density of actions in the training set, giving higher importance to less frequent actions.

\section{Agent}
\label{sec:agent}

\subsection{Input representation}
The input \( s \) to the agent's policy is a six-channel 192x192 image produced by a module in CARLA simulator that has access to the city map. 
This image represents the mid-level bird's-eye view of the vehicle's current position. 
In this work, the BEV contains three channels, the road, the route and the lane, and also
three more channels representing the history of the traffic lights at time steps -16, -9 and -1.

Additionally, the agent receives the current speed of the vehicle and the last values of the policy actuators (previous acceleration and steering). These additional inputs are incorporated further down in the network, specifically in the first fully connected layer.

\subsection{Output Representation}
The vehicle in CARLA has three actuators: steering \([-1, 1]\), throttle \([0, 1]\), and brake \([0, 1]\). Our agent's action space is \(\mathbf{a} \in [-1, 1]^2\), where the two components correspond to steering and acceleration. Braking is implied when acceleration is negative, ensuring the agent cannot brake and accelerate simultaneously \cite{Petrazzini2021}.

Instead of using the Gaussian distribution for the policy's actions, we employ the Beta distribution \(\mathcal{B}(\alpha, \beta)\). The Beta distribution's bounded support is suitable for real-world applications like autonomous driving, where actions are naturally constrained \cite{Petrazzini2021}. This approach avoids the need for clipping or squashing inputs to enforce constraints, allowing explicit computation of the policy loss \(\mathcal{L}_P\).

The Beta distribution also handles extreme driving scenarios, such as sharp turns and sudden braking, effectively. The parameters \(\alpha\) and \(\beta\), which are outputs of the policy neural network \(\pi_\theta\), control the shape of the distribution, enabling the policy to adapt to these critical behaviors.

\subsection{Network architecture}

\begin{figure*}
  \includegraphics[width=\textwidth]{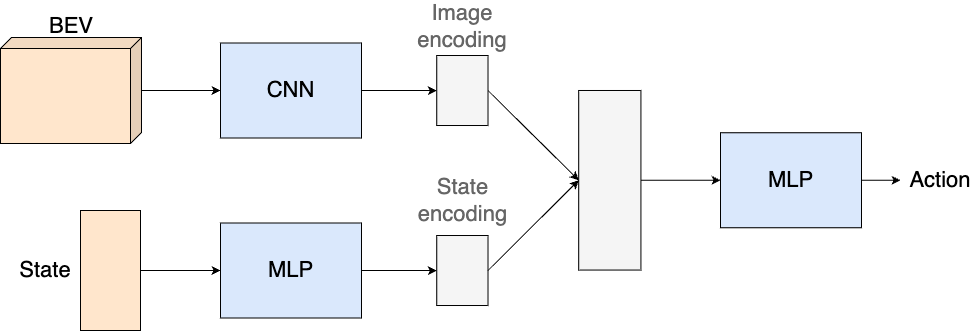}
  \caption{Network architecture. The BEV (state) input is processed by a CNN (MLP) to create an image (state) encoding. Both image and state encodings are concatenated and fed into a final MLP that outputs the actions.}
  \label{fig:architecture}
\end{figure*}

In this work, we train a network architecture, shown in Fig.~\ref{fig:architecture}, to learn to control an autonomous vehicle, using steering and acceleration actuators, based on the current BEV and some state variables as inputs.
The Convolutionary Neural Network (CNN) maps the BEV into an embedding vector of size 256. 
Similarly, the scalar state variables are embedded into a vector of size 256 by a standard MLP (multi-layer perceptron layer). Together they form the current representation of the environment the agent faces. This concatenated embedding vector is processed by another MLP to map to the desired action. 

\begin{figure*}
\centering
\includegraphics[width=0.75\textwidth]{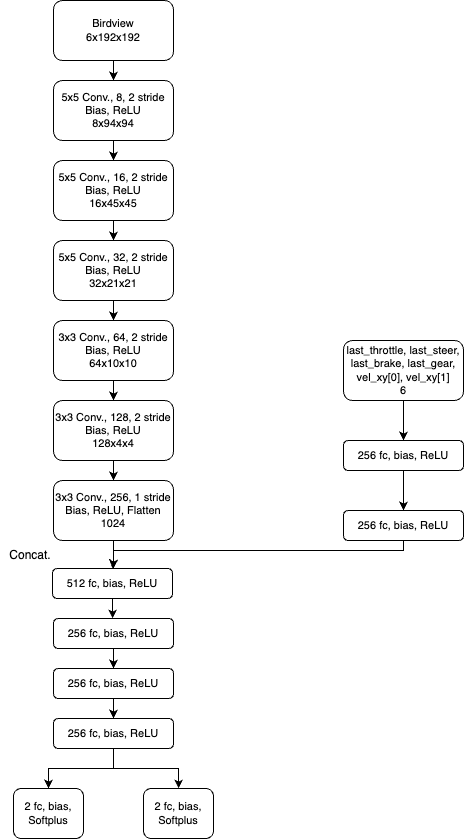}
  \caption{Detailed network architecture. The BEV input consists of 6 channels of 192x192 resolution, while the State input correponds to the last executed actions, and the current speed measurements. The final policy output corresponds to the two parameters $\alpha$ and $\beta$ of the Beta distribution for both acceleration and steering (thus, the two heads). }
  \label{fig:detail_architecture}
\end{figure*}

Fig.~\ref{fig:detail_architecture} shows the detailed network architecture.

\section{Experiments}
\label{sec:experiments}

In this section, the proposed BC with weigthed loss function is compared with the standard BC approach for autonomous vehicle navigation tasks in urban scenarios using the CARLA simulator.

\subsection{Dataset Generation}

The environment and trajectories are sourced from the CARLA Leaderboard evaluation platform \cite{leaderboard_2020}. Specifically, the \textit{town01} environment from this platform, along with ten predefined trajectories, is used to create the expert dataset.

The expert dataset is generated using a deterministic agent that navigates with a dense point trajectory and a classic PID controller \cite{chen2019learning}. 
The dense point trajectory offers numerous points at a high resolution, while a sparse point trajectory has significantly fewer points, giving the agent only a general direction. Therefore, the dense point trajectory is employed to generate the expert's training data, whereas the sparse point trajectory is used by the agent for broader guidance. Thus, the agent has no access to the dense point trajectory.
Besides, the expert breaks at the detection of red traffic lights and accelerates when the light turns back to green.

\begin{figure}[tb]
  \centering
  \subfigure[\textit{town01}]{
  {\includegraphics[scale=0.32]{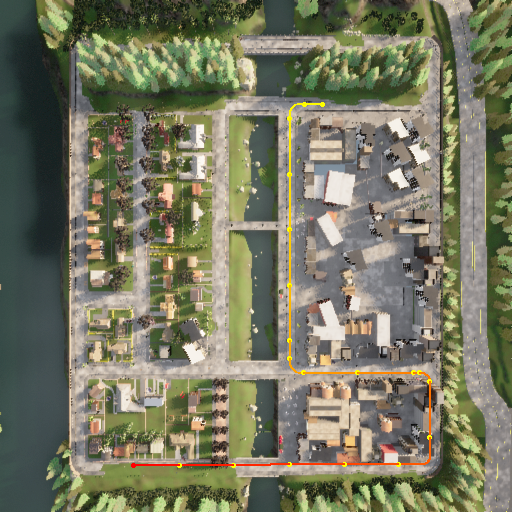}}
  }
  \subfigure[\textit{town02}]{
  {\includegraphics[scale=0.32]{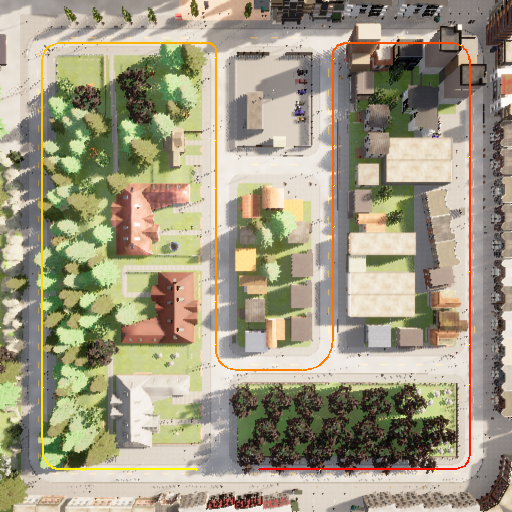}}
  }
  \caption{
  (a) The \textit{town01}  environment of the agent includes one of the routes used by the expert to collect data. This highlighted path measures 740 meters, featuring 20 points in the sparse trajectory (indicated by yellow dots) and 762 points in the dense point trajectory (not shown).
  (b) The \textit{town02} is used to evaluate the trained agent in a new environment.
  } 
  \label{fig:town01}
\end{figure}

In Fig.~\ref{fig:town01}, 
one of the 10 routes executed by the expert to create the labeled training set of demonstrations is illustrated. The line that starts in yellow and ends in red represents the desired trajectory (not visible to the agent as it is). The sparse trajectory appears as yellow dots, generated every 50 meters or when the vehicle is about to change its movement (e.g., from \textit{straight} to \textit{turn} and vice versa).

The ten trajectories in the expert set were recorded at a rate of 10 hertz, producing 10 observation-action pairs per second. For the shortest route of 1480 samples (average route of 2129 samples), this equates to 2.5 minutes (3.5 minutes) of simulated driving. Altogether, the ten trajectories resulted in 21,287 labeled samples (30 GB of uncompressed data). The entire set corresponds to approximately 36 minutes or 8 km of driving.

\subsection{Training and Inference}
The network shown in Fig.~\ref{fig:detail_architecture} is trained with 8 routes from the expert dataset of \textit{town01} environment, where the agent must also respect traffic lights.
The first agent (Weighted BC) weighs the samples according to the density of the actions using (\ref{eq:weighted_loss}) as loss function, whereas the second agent (BC) considers all actions with same weight, i.e., it removes the higher importance of less frequent actions, using (\ref{eq:bcloss_car}) as the conventional BC loss function.

\subsection{Evaluation}

\begin{figure}[tb!]
  \centering
  {\includegraphics[scale=0.5]{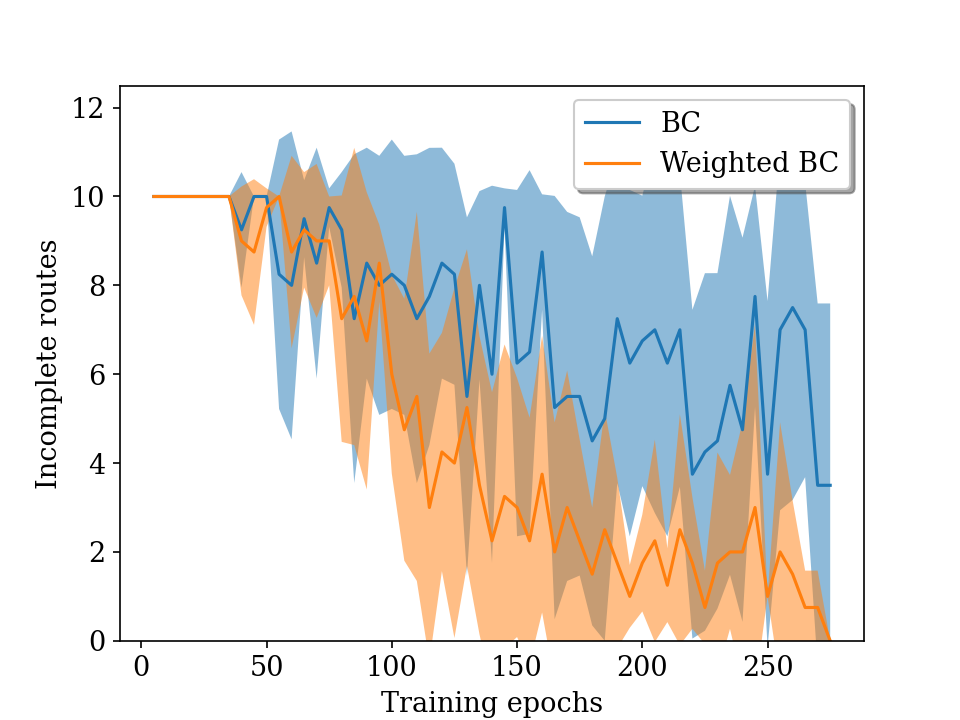}}
  \caption{
  Evaluation of both BC policies (weighted and regular) throughout its learning epochs. 
  At every eight training epochs, both BC and Weighted BC agents are evaluated in all ten routes of \textit{town01}.
  The vertical axis presents the number of incomplete routes. 
  Four runs were executed for each agent, with mean and standard deviation shown by the solid line and the shaded area, respectively.
  In the beginning, both agents commit infractions in all routes and, thus, the number of incomplete routes is ten since one infraction ends the route uncompleted. The Weighted BC agent is able to achieve zero infractions by the end of training, unlike the BC agent.
  } 
  \label{fig:infractions}
\end{figure}

Although training is offline in \textit{town01} environment, we still evaluate both BC and weighted BC agents as their model's parameters are updated. Every eight training epochs, both agents are evaluated using all ten routes from the same training town (Fig. \ref{fig:infractions}). As four random runs are executed for each agent, the mean and standard deviation of the number of incomplete routes is plot in Fig. \ref{fig:infractions}. One incomplete route means that the vehicle committed an infraction before completing the route. As the total number of routes is ten for \textit{town01}, both BC and Weighted BC start at 10, the minimum performance. As training evolves, both agents improve the performance by committing less infractions, such as going out of the correct lane or passing a red light, but the Weighted BC agent attains the minimum error faster, i.e., it can complete all routes without any infractions and respecting traffic lights.

\begin{figure}[tb!]
  \centering
  {\includegraphics[scale=0.7]{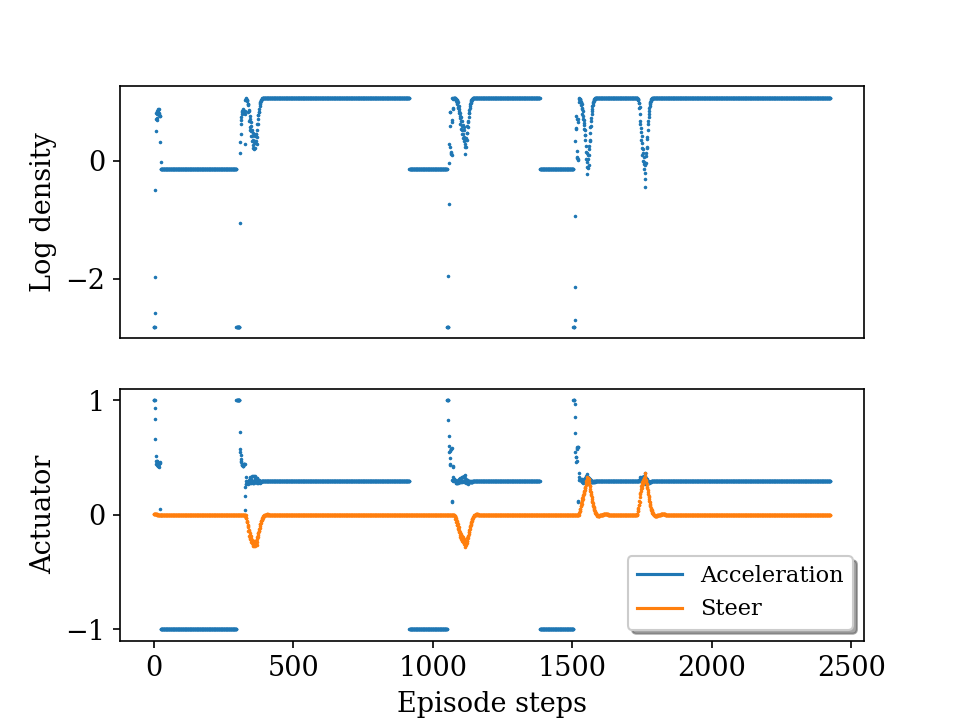}}
  \caption{
  Logarithm of the density of actions, given by KDE, during navigation of a route in \textit{town1}, shown in Fig.~\ref{fig:town01}. The expert encounters three red traffic lights, during which it breaks (acceleration is $-1$), and subsequently accelerates as soon as it turns green. Two left turns and two right turns can be observed by examining the steer actuator of the expert.
  } 
  \label{fig:kde}
\end{figure}

To understand what the action's density given by KDE and, consequently, the weights in the BC loss function, look like for the first route in \textit{town01} (shown in Fig.~\ref{fig:town01}), we record both actuators's values, acceleration and steering, from our expert as it follows the route as well as the corresponding logarithm of the action's density for each timestep of the route.
These values are plot in Fig.~\ref{fig:kde}, where we can see that different actions will produce different weights in the loss function since some of them are less frequent than others in the expert dataset. The highest action density corresponds to going straight with constant acceleration and zero steering, for instance, between timesteps 470 and 900. The least frequent action happens after a traffic light turns from red to green, which corresponds to a higher, but shorter, acceleration period.

The driving performance for both agents as well as the expert is presented in Table 1.
The understanding of the $\uparrow$ or $\downarrow$ in the table is as follows. For instance, the lower ($\downarrow$) the Red Light Infraction, which is \#/Km or the number of infractions per kilometer, the better. In the same vein, the higher the Infraction Penalty (or the Driving score), the better is the performance.
The expert has maximum driving score, being able to complete all routes with no red light infraction in both \textit{town01} and \textit{town02}. The number of crossed traffic lights is just a non-performance metric, just counting the times that the vehicle cross them.
An equivalent result is achieved by Weighted BC in \textit{town01}, the training environment, but not by regular BC, which presents a mean driving score of 81.
In \textit{town02}, a test environment, Weighted BC performs significantly better than BC.
Videos with the agent controlling the vehicle in CARLA are available at
\href{https://sites.google.com/view/weighted-bc}{\texttt{https://sites.google.com/view/weighted-bc}}, comparing both BC and Weighted BC agents.

\begin{table*}[tb!]
\begin{threeparttable}
\centering
\caption{
Driving performance and infraction analysis for different agents on Town01 and Town02. 
}
\begin{tabular}{|p{1.65cm}|p{1.3cm}|p{1.5cm}|p{1.5cm}|p{1.6cm}|p{1.8cm}|p{1.8cm}|}
\hline
\multirow{2}{*}{\textbf{Agent}} & \multirow{2}{*}{\textbf{Town}} & \multicolumn{5}{c|}{\textbf{Metrics}} \\
\cline{3-7}
& & \small{\textbf{Driving Score (\%,$\uparrow$)\tnote{a}}} & \small{\textbf{Route Completion (\%,$\uparrow$)}} & \small{\textbf{Infraction Penalty (\%,$\uparrow$)}} & \small{\textbf{Red Light Infraction (\#/Km,$\downarrow$)}} & \small{\textbf{Crossed traffic lights (\#)}} \\
\hline
\multirow{2}{*}{\textbf{Expert}} & \textit{town01} & 100 $\pm$ 0 & 100 $\pm$ 0 & 100 $\pm$ 0 & 0 $\pm$ 0 & 4.7 $\pm$ 1.9 \\
& \textit{town02} & 100 $\pm$ 0 & 100 $\pm$ 0 & 100 $\pm$ 0 & 0 $\pm$ 0 & 9.0 $\pm$ 2.16 \\
\hline
\multirow{2}{*}{\textbf{BC}} & \textit{town01} & 81 $\pm$ 20 & 83 $\pm$ 18 & 97 $\pm$ 5 & 0.39 $\pm$ 0.68 & 3.78 $\pm$ 1.03 \\
& \textit{town02} & 34 $\pm$ 18 & 53 $\pm$ 18 & 66 $\pm$ 20 & 3.08 $\pm$ 2.45 & 2.79 $\pm$ 2.27 \\
\hline
\textbf{Weighted} & \textit{town01} & 100 $\pm$ 0 & 100 $\pm$ 0 & 100 $\pm$ 0 & 0 $\pm$ 0 & 4.7 $\pm$ 1.9 \\
\textbf{BC} & \textit{town02} & 98 $\pm$ 0.03 & 100 $\pm$ 0 & 98 $\pm$ 0.03 & 0 $\pm$ 0 & 9.0 $\pm$ 2.16 \\
\hline
\end{tabular}

\begin{tablenotes}
\item[a] The \textit{Driving Score}, a metric from CARLA Leaderboard, is calculated by the product of \textit{Route Completion} (the percentage of the route distance completed) and an \textit{Infraction Penalty} which accounts for all violations committed. For instance, if an agent ran two red lights during a route and the penalty for running one red light is 0.7, then the final penalty would be $0.7 \times 0.7 = 0.49$.
\end{tablenotes}
\end{threeparttable}
\label{tab:drivingPerformance}
\end{table*}

\section{Conclusion}
\label{sec:conclusion}

In this work, we proposed a weighted behavior cloning approach that improves agent learning by giving higher weights to underrepresented actions in the expert training set of driving behaviors.
The application consisted of route navigation in two simulated cities in the CARLA realistic simulator for autonomous vehicles, while respecting traffic lights.
As the action space of the expert is not uniformly visited, and the expert training dataset is somewhat limited (only 10 routes in \textit{town01}), a Kernel Density Estimator was employed to infer the action density which, in turn, was converted to a weight per observation-action pair. 
In our investigated scenario, the weighted loss function for BC proved to be crucial to learn the correct behavior, showing significantly better driving performance than regular BC, specially for the test city \textit{town02} not used during training.
Future work include extensions to cities with pedestrians and other vehicles and more advanced behavior cloning approaches such as implicit behavioral cloning \cite{Florence2022} to account for multimodal action distributions.

\begin{credits}

\subsubsection{\discintname}
The authors have no competing interests to declare that are
relevant to the content of this article. 
\end{credits}
\bibliographystyle{splncs04}
\bibliography{biblio}

\end{document}